\pdfoutput=1

\documentclass[11pt]{article}

\usepackage{acl}

\usepackage{times}
\usepackage{latexsym}

\usepackage[T1]{fontenc}

\usepackage[utf8]{inputenc}

\usepackage{microtype}

\usepackage{inconsolata}
\usepackage{color}
\usepackage[noend]{algpseudocode}
\usepackage{bm}
\usepackage{algorithmicx,algorithm}
\usepackage{multirow}
\usepackage{subcaption}
\usepackage{amsfonts}
\usepackage{amsmath}
\usepackage{booktabs}
\usepackage{makecell}
\usepackage{graphicx}

\usepackage{CJKutf8}

\title{Budget-Constrained Tool Learning with Planning}

\author{Yuanhang Zheng$^{1}$, Peng Li$^{2*}$, Ming Yan$^{3}$, Ji Zhang$^{3}$, Fei Huang$^{3}$ and Yang Liu$^{1,2,4}$\thanks{\ \ Corresponding authors: P.Li (\href{mailto:lipeng@air.tsinghua.edu.cn}{\texttt{lipeng@air.tsinghua.}} \href{mailto:lipeng@air.tsinghua.edu.cn}{\texttt{edu.cn}}) and Y.Liu (\href{mailto:liuyang2011@tsinghua.edu.cn}{\texttt{liuyang2011@tsinghua.edu.cn}}).} \\
    $^1$Dept. of Comp. Sci. \& Tech., Institute for AI, Tsinghua University, Beijing, China \\
    $^2$Institute for AI Industry Research (AIR), Tsinghua University, Beijing, China \\
    $^3$Institute of Intelligent Computing, Alibaba Group \\
    $^4$Jiangsu Collaborative Innovation Center for Language Competence, Jiangsu, China
}

\begin{document}
\begin{CJK}{UTF8}{gbsn}

\maketitle
\begin{abstract}
Despite intensive efforts devoted to tool learning, the problem of budget-constrained tool learning, which focuses on resolving user queries within a specific budget constraint, has been widely overlooked. This paper proposes a novel method for budget-constrained tool learning. Our approach involves creating a preferable plan under the budget constraint before utilizing the tools. This plan outlines the feasible tools and the maximum number of times they can be employed, offering a comprehensive overview of the tool learning process for large language models. This allows them to allocate the budget from a broader perspective. To devise the plan without incurring significant extra costs, we suggest initially estimating the usefulness of the candidate tools based on past experience. Subsequently, we employ dynamic programming to formulate the plan. Experimental results demonstrate that our method can be integrated with various tool learning methods, significantly enhancing their effectiveness under strict budget constraints.\footnote{Code can be found at \url{https://github.com/THUNLP-MT/BTP}.}
\end{abstract}

\section{Introduction}

Tool learning~\citep{schick2023toolformer,yao2023react,qin2023toolllm}, which uses external tools to extend the capability of large language models (LLMs), has achieved remarkable results on various types of tasks. For example, with the aid of external tools, LLMs may solve difficult mathematical problems with higher accuracy~\citep{cobbe2021training,gao2023pal}, handle multimodal information~\citep{shen2023hugginggpt,lu2023chameleon}, or interact with real-world applications~\citep{song2023restgpt,gur2023realworld}. To better resolve more complex user queries, previous studies have proposed different tool learning methods which allow the LLM to use multiple tools~\citep{chen2023chatcot,yao2023react,paranjape2023art,shen2023hugginggpt,qin2023toolllm}.

\begin{table}
\centering
\small
\setlength{\tabcolsep}{4pt}
\begin{tabular}{lcccccc}
\toprule
\multirow{4}{*}{\textbf{Method}} & \multicolumn{6}{c}{\textbf{Budget Constraint}} \\\cmidrule{2-7}
& \multicolumn{3}{c}{$+\infty$} & \multicolumn{3}{c}{20} \\\cmidrule(lr){2-4}\cmidrule(lr){5-7}
& PR$\uparrow$ & PBC$\uparrow$ & AC$\downarrow$ & PR$\uparrow$ & PBC$\uparrow$ & AC$\downarrow$ \\\midrule
ReAct & 44.0 & 44.0 & 15.4 & \textbf{44.0} & 34.1 & 15.4 \\
\ +BTP (\emph{Ours}) & \textbf{46.3} & \textbf{46.3} & \textbf{9.0} & 43.7 & \textbf{43.7} & \textbf{\hphantom{0}6.9} \\\midrule
DFSDT & 63.8 & 63.8 & 78.3 & 63.8 & 28.8 & 78.3 \\
\ +BTP (\emph{Ours}) & \textbf{66.1} & \textbf{66.1} & \textbf{12.5} & \textbf{64.5} & \textbf{64.5} & \textbf{\hphantom{0}9.2} \\
 \midrule
 ToT-DFS & 61.6 & 61.6 & 51.4 & 61.6 & 10.2 & 51.4 \\
 \ +BTP (\emph{Ours}) & \textbf{65.0} & \textbf{65.0} & \textbf{15.8} & \textbf{64.1} & \textbf{64.1} & \textbf{10.8} \\
\bottomrule
\end{tabular}
\caption{Comparison of \textbf{Pass Rate (PR)}, \textbf{Pass rate under Budget Constraint (PBC)} and \textbf{Average Cost (AC)} on the ToolBench~\citep{qin2023toolllm} dataset. Our proposed \textbf{B}udget-Constrained \textbf{T}ool Learning with \textbf{P}lanning (BTP) reduces the cost of tool learning and reaches competitive Pass Rate, significantly improving the performance under a strict budget constraint.}
\label{tab:exp1}
\end{table}

\begin{figure*}[t]
\begin{center}
\includegraphics[scale=0.42]{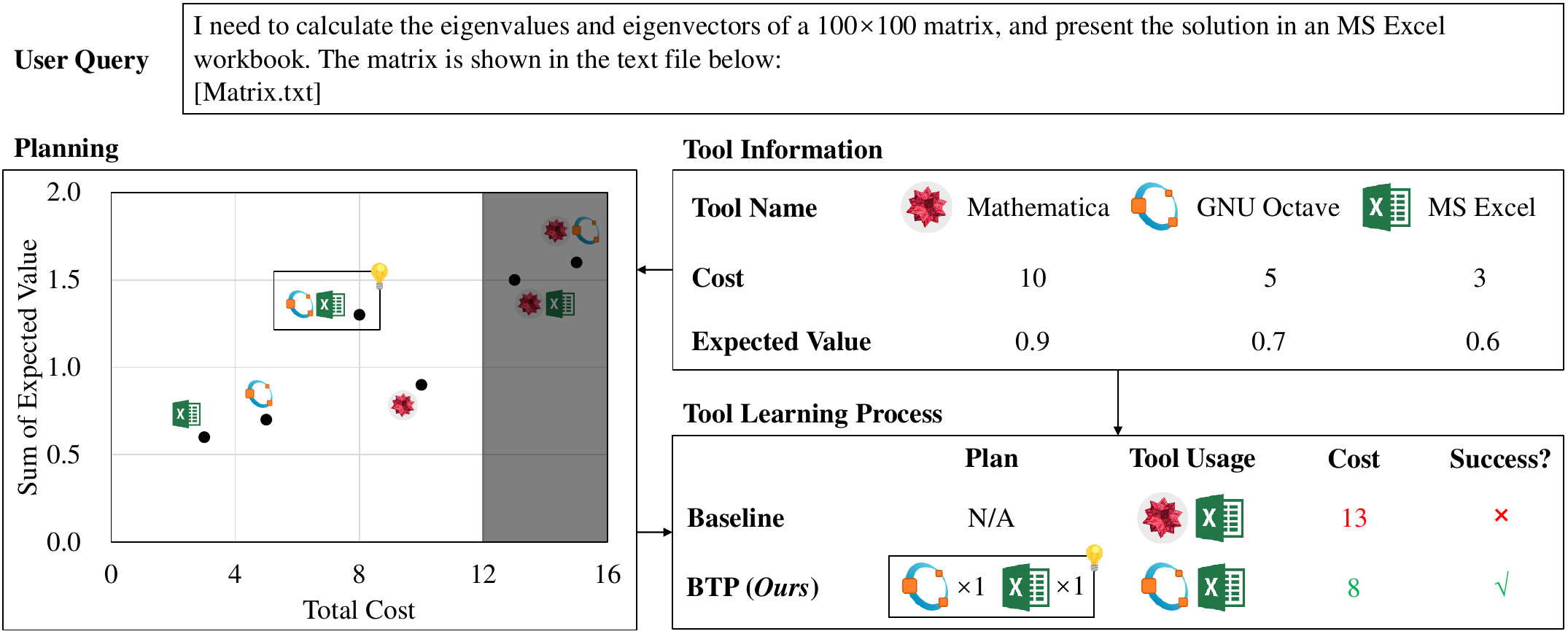} 
\caption{Example of budget-constrained tool learning. In this example, the baseline fails to resolve the user query within the budget constraint. Our proposed BTP makes a preferable plan of tool usage before using the tools, which may help resolve the user query under the budget constraint. ``Expected Value'' measures how valuable the candidate tool is for resolving the given query, which is estimated based on the past experience of tool learning. The shaded area in the ``Planning'' part means that the candidate plans in the area exceed the budget constraint.}
\label{fig:idea}
\end{center}
\end{figure*}

Despite their effectiveness, existing methods often overlook a critical aspect: 
utilizing tools incurs expenses, such as money and time, and users may have (implicit) budget constraints in real-world scenarios. 
Without meticulous management, the costs can rapidly exceed the acceptable threshold of a user. In scenarios where the expenses overshadow the benefits, even if the tool successfully addresses the intended problem, users may still perceive the solution as unsatisfactory due to the disproportionate incurred costs.
Table~\ref{tab:exp1} presents preliminary results for three strong tool learning approaches, ReAct~\citep{yao2023react}, DFSDT~\citep{qin2023toolllm}, and ToT-DFS~\citep{yao2023tree}, evaluated on the ToolBench~\citep{qin2023toolllm} dataset, with success defined strictly in terms of problem resolution within budget constraints. It is evident that all three methods experience significant declines in performance as the budget constraint tightens from unlimited to 20.
Consequently, we argue that more efforts are worth devoting to the problem of \emph{budget-constrained tool learning}, which aims at resolving user queries within a given budget constraint.

One fundamental challenge in budget-constrained tool learning is the allocation of the budget. However, determining the optimal budget allocation is not straightforward without a comprehensive understanding of the entire tool learning process. Taking Figure~\ref{fig:idea} as an example, the user query may be addressed using either Mathematica and MS Excel, or GNU Octave and MS Excel. While Mathematica, a commercial software, may provide a more precise solution than the open-source GNU Octave, we may incorrectly choose Mathematica without knowing the additional requirement for MS Excel. Unfortunately, acquiring such a comprehensive view is difficult. A simplistic approach is employing trial-and-error methods. However, each attempt introduces additional costs, further complicating adherence to the budget constraint.

To this end, we propose \textbf{B}udget-Constrained \textbf{T}ool Learning with \textbf{P}lanning (BTP), a novel method for budget-constrained tool learning, which finds a preferable plan under the budget constraint before using the tools (Figure~\ref{fig:idea}). First, for each candidate tool, we estimate its expected value and frequency constraint based on the past experience. The expected value measures how valuable the tool is for resolving the given query, while the frequency constraint limits the maximum number of times it can be used. Then, we try to find a plan that maximizes the sum of the expected value under the budget constraint. A plan specifies the viable tools and how many times each of them can be used. We regard this as a knapsack problem and use dynamic programming to resolve it. Finally, we apply the plan during the tool learning process. As shown in Table~\ref{tab:exp1}, our proposed method can be combined with different tool learning methods and improve their performance under a strict budget constraint.

\section{Methodology}

In this section, we first introduce a formulation of budget-constrained tool learning (Section~\ref{sec:budget_constrained_tool_learning}). Then, we describe our proposed BTP in detail. Before using the tools, we try to find a preferable plan under the budget constraint (Section~\ref{sec:tool_plan}). Then, we apply the plan to the process of tool learning (Section~\ref{sec:apply_plan}). Moreover, we introduce a blacklist mechanism to further reduce the cost of tool learning (Section~\ref{sec:cache}).

\subsection{Budget-Constrained Tool Learning}
\label{sec:budget_constrained_tool_learning}

Existing tool learning methods can be roughly divided into two lines. The first line finetunes the LLMs to make them capable of using tools without the documentation of the tools~\cite{schick2023toolformer,hao2023toolkengpt}, while the second line lets the LLM select tools according to the documentation of the candidate tools~\cite{shen2023hugginggpt,hsieh2023tool,qin2023toolllm}. In this paper, we mainly focus on the latter line, since it can better support closed-source LLMs, including those which do not support fine-tuning. Typically, the tool learning process following this line is as follows: First, a set of $n$ candidate tools $\mathcal{T}=\{t_1, \dots, t_i, \dots, t_n\}$ are retrieved according to the user query using a retrieval model. Then an LLM is leveraged to select and invoke tools iteratively until a stop condition is met. Assume that this process runs for $N$ times, we denote the tools invoked as $\mathcal{T}_{\mathrm{I}}=\{t_{a_1},\cdots,t_{a_j},\cdots,t_{a_N}\}$. Note that a tool may be invoked multiple times. Finally, the LLM produces the final output based on the user query and the results returned by the executed tools.

To conduct budget-constrained tool learning, we first propose a simplified and approximate but useful mathematical formulation for it. 
Formally, for any integer $i \in [1,n]$, we use $c_i$ to represent the cost of using the tool $t_i$ once.\footnote{For simplicity, we also call $c_i$ as ``the cost of $t_i$'' in the remaining part of this paper.} In reality, this cost may be the estimated total cost spent on the tool in different aspects, including the money spent on using the tool and the LLM, as well as the consumed time for using the tool and the LLM:
\begin{equation}
    c_i=c_{i,\mathrm{tool}}+c_{\mathrm{LLM}}+T_i,
\end{equation}
where $c_{i,\mathrm{tool}}$ represents the estimated cost of using the tool $t_i$ itself, $c_{\mathrm{LLM}}$ is the estimated cost of using the API of the LLM once, and $T_i$ is the estimated consumed time for using the tool and the LLM. We implicitly assume that all costs involved can be quantified using the same unit of measurement. For example, we can apply a conversion function to translate time costs into monetary terms.

Moreover, we use $c_s$ to represent the extra cost produced by the other factors such as the system prompt and the user prompt, and $B$ to represent the budget constraint. Then, \emph{the target of budget-constrained tool learning is to resolve a user query under the budget constraint:}
\begin{equation}
c_s+\sum\limits_j c_{a_j} \leq B.
\label{eq:constraint}
\end{equation}

In practice, $c_s$ and $c_i$ are influenced by numerous factors, complicating the design of a feasible algorithm. To simplify the problem and make it manageable, we introduce an assumption that both $c_s$ and $c_i$ can be estimated in advance. The assumption does not significantly limit the generality of our approach. On the one hand, both of them can be estimated using their historical average values. On the other hand, $c_s$ can be more precisely estimated based on the price of the API of the LLM, the lengths of the system prompt and the user prompt for the current user query.

Currently, we do not take the cost arising from the external environments into consideration. For example, when the LLM uses an online shopping tool to buy a book for the user, the price of the book also consumes the overall budget and is difficult to be estimated in advance. Addressing this aspect is designated for future research.

\subsection{Tool Usage Planning}
\label{sec:tool_plan}
In this work, we introduce a tool usage plan to offer an estimated overview of the tools likely to be utilized by the LLM, aiding in more effective budget allocation. The plan outlines the set of available tools and the maximum number of times each tool can be used.

However, it is difficult to precisely calculate how many times each candidate tool should be used for resolving a given user query. On the one hand, it is not trivial to estimate whether a candidate tool can provide valuable information for resolving the user query without using the tool. On the other hand, calculating the costs of every possible plan of tool usage for a given user query is unacceptable in terms of time complexity. Thus, we propose an approximate planning method for budget-constrained tool learning. In this planning method, we estimate whether the candidate tools can provide valuable information based on the past experience, and use dynamic programming to estimate how many times each candidate tool should be used.

Specifically, we first estimate the expected value of each candidate tool, which measures how valuable the results returned by the candidate tool are. In this work, we leverage the past experience of tool learning to estimate the expected value.
Formally, the past experience $\mathcal{E}=\{U_1,\dots,U_k,\dots,U_m\}$ consists of a number of tool usages $U_k$ which happened in the past.
Each tool usage $U_k=\langle q, d(t), p, r\rangle$ consists of a user query $q$, the documentation $d(t)$ of the used tool $t$, the input parameters $p$ and the returned result $r$.
We define a binary score function $\mathrm{score}(\cdot)$ to judge the usefulness of $r$ for resolving $q$, where $\mathrm{score}(U_k)=0$ or $1$ means that $r$ is unhelpful or helpful, respectively. In practice, $\mathrm{score}(\cdot)$ is implemented as a LongFormer~\citep{beltagy2020longformer} based classifier.

Then, for each candidate tool $t_i$, we fetch all tool usages $U_k$ related to $t_i$ in the past experience $\mathcal{E}$ to construct a set of tool usages $\mathcal{E}_{t_i}$. We calculate the weighted average score of all $U_k \in \mathcal{E}_{t_i}$ as the expected value of $t_i$ for the user query $q_u$:
\begin{equation}
v(q_u,t_i)=\frac{\sum\limits_{U_k \in \mathcal{E}_{t_i}} \mathrm{exp}(\mathrm{sim}(q_u,q))\mathrm{score}(U_{k})}{\sum\limits_{U_k \in \mathcal{E}_{t_i}} \mathrm{exp}(\mathrm{sim}(q_u,q))},
\end{equation}

\noindent where $\mathrm{sim}(q_u,q)$ is the similarity between $q_u$ and $q$, which can be calculated using the retrieval model.

We also notice that repeatedly using the same tool too many times may be suboptimal within the budget constraint, since we expect that more diverse information can be obtained if different tools are used. Thus, we set an estimated frequency constraint $\tilde{F}(q_u,t_i)$ for each candidate tool $t_i$ and expect that $t_i$ should be used no more than $\tilde{F}(q_u,t_i)$ times. Specifically, if a tool frequently returns useless or erroneous messages, we should prevent the LLM from using it to reduce the cost of tool learning. Thus, we set a threshold $\tau$ and set $\tilde{F}(q_u,t_i)=0$ if $v(q_u,t_i)<\tau$ to filter out the useless tools based on the past experience. Otherwise, for each query $q$ in the past experience where $t_i$ is used in the process of resolving $q$, we count the number of times $F(q,t_i)$ for which $t_i$ is used, and then calculate the weighted average of $F(q,t_i)$ as the estimated frequency constraint $\tilde{F}(q_u,t_i)$:
\begin{equation}
\tilde{F}(q_u,t_i)=\frac{\sum\limits_{q} \mathrm{exp}(\mathrm{sim}(q_u,q))F(q,t_i)}{\sum\limits_{q} \mathrm{exp}(\mathrm{sim}(q_u,q))}.
\end{equation}

Finally, we regard the planning process of budget-constrained tool learning as a knapsack problem and use dynamic programming to find a preferable plan of tool usage within the budget constraint. Formally, for each candidate tool $t_i$, the plan specifies its corresponding frequency $f(q_u,t_i)$, which means that we expect the LLM to use $t_i$ for at most $f(q_u,t_i)$ times during the tool learning process. Since we expect that the returned results should contain valuable information as much as possible, the plan is determined by maximizing the sum of the expected value:
\begin{equation}
V=\sum\limits_{i=1}^{n} f(q_u,t_i)v(q_u,t_i)
\end{equation}
under the budget and frequency constraints:
\begin{align}
\sum\limits_{i=1}^{n} f(q_u,t_i)c_i &\leq R, \\
f(q_u,t_i) &\leq \tilde{F}(q_u,t_i),
\end{align}
where $R=B-c_s$. For simplicity, we assume that $c_i$ and $R$ are positive integers.\footnote{If $c_i$ and $R$ are real numbers, we can use their approximate values and change them into integers. Theoretically, this approximation can achieve any precision. For the detailed proof, please refer to Appendix~\ref{app:proof_approximation}.} The details are shown in Algorithm~\ref{alg:dp}.

The process of finding the plan also introduces extra cost. Fortunately, the cost is a near-constant value. Thus, we can simply extend $c_s$ to include this extra cost and keep the algorithm untouched.

Moreover, it is worth noting that our proposed dynamic programming method can be regarded as a searching algorithm.
Searching-based methods are widely used in previously proposed tool learning methods such as DFSDT~\citep{qin2023toolllm} and ToT-DFS~\citep{yao2023tree}. However, these methods are not feasible for searching tool usage plans as they need to invoke the tools during searching, which consume the budget.

\begin{algorithm}[t]
\caption{Tool Usage Planning\label{alg:dp}}
{\bf Input:}
budget constraint $B$; estimated cost for system prompt and user prompt $c_s$; estimated cost $c_i$, expected value $v(q_u,t_i)$ and frequency constraint $\tilde{F}(q_u,t_i)$ for each tool in $\mathcal{T}=\{t_1,\dots,t_i,\dots,t_n\}$\\
{\bf Output:}
frequency $f(q_u,t_i)$
\begin{algorithmic}[1]
\State $R \leftarrow B-c_s$
\For {$j \leftarrow 0$ to $R$}
    \State $V_{0,j} \leftarrow 0$
\EndFor
\For {$i \leftarrow 1$ to $n$}
    \For {$j \leftarrow 0$ to $R$}
        \State $V_{i,j} \leftarrow V_{i-1,j}$
        \For {$k \leftarrow 1$ to $\lfloor \tilde{F}(q_u,t_i) \rfloor$}
            \If {$j \geq kc(t_i)$}
                \State $V_{i,j} \leftarrow \mathrm{max}(V_{i,j},V_{i,j-kc(t_i)}+kv(q_u,t_i))$
            \EndIf
        \EndFor
    \EndFor
\EndFor
\State $V_{max} \leftarrow 0$
\For {$j \leftarrow 0$ to $R$}
    \State $V_{max} \leftarrow \mathrm{max}(V_{max}, V_{n,j})$
\EndFor
\State Trace back the dynamic programming process to obtain the frequency $f(q_u,t_i)$
\end{algorithmic}
\end{algorithm}

\begin{algorithm}[t]
\caption{Budget-Constrained Tool Learning with Plan\label{alg:tool_learning_budget_plan}}
{\bf Input:}
user query $q_u$, set of candidate tools $\mathcal{T}=\{t_1,\dots,t_i,\dots,t_n\}$, frequency $f(q_u,t_i)$ for each tool $t_i$\\
{\bf Output:}
returned result $y$
\begin{algorithmic}[1]
\For {each $t_i \in \mathcal{T}$}
    \If {$f(q_u,t_i)=0$}
        \State $\mathcal{T} \leftarrow \mathcal{T}\setminus\{t_i\}$
    \EndIf
\EndFor
\State $j \leftarrow 0$
\While {True}
    \State $j \leftarrow j+1$
    \State $t_{a_j}, p_j \leftarrow \mathrm{UseTool}(q_u,\mathcal{T},r_1,\dots,r_{j-1})$ 
    \Statex \Comment{{\color{gray}\it \hspace{-0.4em} Select a tool and determine its parameter}}
    \State $r_j \leftarrow t_{a_j}(p_j)$ \Comment{{\color{gray}\it Invoke the tool to get result}}
    \State $f(q_u,t_{a_j}) \leftarrow f(q_u,t_{a_j})-1$
    \If {$f(q_u,t_{a_j})=0$}
        \State $\mathcal{T} \leftarrow \mathcal{T}\setminus\{t_{a_j}\}$
    \EndIf
    \If {$\mathcal{T}=\emptyset$ or $\mathrm{IsSufficient}(q_u,r_1,\dots,r_j)$} 
    \Comment{{\color{gray}\it If there are no available tools in $\mathcal{T}$ or the returned results are sufficient for resolving the user query}}
        \State $y \leftarrow \mathrm{Summarize}(q_u,r_1,\dots,r_j)$ 
        \Statex \Comment{{\color{gray}\it Summarize to get the returned result}}
        \State \Return $y$
    \EndIf
\EndWhile
\end{algorithmic}
\end{algorithm}

\subsection{Applying Plan to Tool Learning}
\label{sec:apply_plan}

After we obtain the frequency $f(q_u,t_i)$ for each candidate tool $t_i$, we apply the plan to the tool learning process. Specifically, before the LLM uses tools, we remove all the tools $t_i$ which satisfy $f(q_u,t_i)=0$ from $\mathcal{T}$ since they should not be used in the process. Then, after each step $j$ when the LLM uses the tool $t_{a_j}$, we reduce the frequency $f(q_u,t_{a_j})$ by 1. Once this frequency reaches zero, we remove $t_{a_j}$ from $\mathcal{T}$, which indicates that $t_{a_j}$ should be no more used. We add extra information in the input context to inform the LLM that the tool $t_{a_j}$ is forbidden to use. If the LLM occasionally uses the forbidden tools, we prevent the LLM from the actual action of the tool usage and return an error message.
Finally, if $\mathcal{T}$ becomes empty or the returned results are sufficient to draw a conclusion, we make a summary about all the returned results. The aforementioned process is shown in Algorithm~\ref{alg:tool_learning_budget_plan}.

\subsection{Blacklist Mechanism}
\label{sec:cache}

To further reduce the cost of tool learning, we introduce a blacklist mechanism to reduce the cost of repeatedly using unhelpful tools. During the tool learning process, we temporarily build the blacklist according to the judgment about the returned results. Specifically, after the LLM uses a tool and receives the returned result, we use the classifier introduced in Section~\ref{sec:tool_plan} to judge whether the returned result is helpful in resolving the user query.\footnote{The cost of the classifier can also be added to $c_i$.} If a tool returns unhelpful results, we list it into the blacklist and forbid the LLM from using it. Specifically, we inform the LLM that the blacklisted tools are forbidden to use by giving the blacklisted tools in the input context during the tool learning process. If the LLM occasionally uses a blacklisted tool, we prevent the LLM from the actual action of the tool usage and return an error message.

\begin{table*}
\centering
\small
\setlength{\tabcolsep}{3.5pt}
\begin{tabular}{lcccccccccccccc}
\toprule
\multirow{2.5}{*}{\textbf{Method}} & \multicolumn{2}{c}{\textbf{I1-Inst}} & \multicolumn{2}{c}{\textbf{I1-Tool}} & \multicolumn{2}{c}{\textbf{I1-Cat}} & \multicolumn{2}{c}{\textbf{I2-Inst}} & \multicolumn{2}{c}{\textbf{I2-Cat}} & \multicolumn{2}{c}{\textbf{I3-Inst}} & \multicolumn{2}{c}{\textbf{Average}}\\\cmidrule(lr){2-3}\cmidrule(lr){4-5}\cmidrule(lr){6-7}\cmidrule(lr){8-9}\cmidrule(lr){10-11}\cmidrule(lr){12-13}\cmidrule(lr){14-15}
& PBC$\uparrow$ & AC$\downarrow$ & PBC$\uparrow$ & AC$\downarrow$ & PBC$\uparrow$ & AC$\downarrow$ & PBC$\uparrow$ & AC$\downarrow$ & PBC$\uparrow$ & AC$\downarrow$ & PBC$\uparrow$ & AC$\downarrow$ & PBC$\uparrow$ & AC$\downarrow$ \\\midrule
ReAct & 38.5 & 15.8 & 37.0 & 16.3 & 38.5 & 15.0 & 33.5 & 15.8 & 35.0 & 17.0 & 22.0 & 12.6 & 34.1 & 15.4 \\
\ +BTP (\emph{Ours}) & \textbf{43.5} & \hphantom{0}\textbf{7.6} & \textbf{48.0} & \hphantom{0}\textbf{7.8} & \textbf{49.5} & \hphantom{0}\textbf{7.2} & \textbf{41.5} & \textbf{7.0} & \textbf{49.5} & \hphantom{0}\textbf{7.1} & \textbf{30.0} & \hphantom{0}\textbf{4.7} & \textbf{43.7} & \hphantom{0}\textbf{6.9} \\\midrule
ReAct+Prompt & 35.5 & 16.7 & 39.5 & 17.1 & 43.0 & 14.4 & 37.5 & 14.9 & 39.5 & 16.3 & 24.0 & 14.9 & 36.5 & 15.7 \\
\ +BTP (\emph{Ours}) & \textbf{47.0} & \hphantom{0}\textbf{8.7} & \textbf{44.0} & \hphantom{0}\textbf{8.3} & \textbf{50.5} & \hphantom{0}\textbf{7.8} & \textbf{47.0} & \hphantom{0}\textbf{7.7} & \textbf{54.0} & \hphantom{0}\textbf{8.0} & \textbf{32.0} & \hphantom{0}\textbf{5.2} & \textbf{45.8} & \hphantom{0}\textbf{7.6} \\\midrule
DFSDT & 34.0 & 77.5 & 29.5 & 75.1 & 34.5 & 70.1 & 28.0 & 81.6 & 28.5 & 79.4 & 18.0 & 85.8 & 28.8 & 78.3 \\
\ +BTP (\emph{Ours}) & \textbf{58.0} & \hphantom{0}\textbf{9.6} & \textbf{58.0} & \textbf{10.3} & \textbf{63.5} & \hphantom{0}\textbf{9.5} & \textbf{77.0} & \hphantom{0}\textbf{9.5} & \textbf{66.5} & \hphantom{0}\textbf{9.8} & \textbf{64.0} & \hphantom{0}\textbf{6.4} & \textbf{64.5} & \hphantom{0}\textbf{9.2} \\\midrule
DFSDT+Prompt & 31.0 & 69.5 & 29.5 & 84.8 & 36.5 & 61.2 & 28.5 & 68.4 & 31.0 & 74.4 & 17.0 & 81.3 & 29.0 & 73.3 \\
\ +BTP (\emph{Ours}) & \textbf{55.0} & \textbf{10.0} & \textbf{58.0} & \textbf{10.1} & \textbf{63.0} & \hphantom{0}\textbf{9.2} & \textbf{77.5} & \hphantom{0}\textbf{9.5} & \textbf{65.5} & \textbf{10.0} & \textbf{64.0} & \hphantom{0}\textbf{6.1} & \textbf{63.8} & \hphantom{0}\textbf{9.2} \\\midrule
ToT-DFS & 11.5 & 52.4 & 10.5 & 52.4 & 11.0 & 48.8 & \hphantom{0}7.0 & 52.4 & 11.0 & 53.3 & 10.0 & 48.9 & 10.2 & 51.4 \\
\ +BTP (\emph{Ours}) & \textbf{55.5} & \textbf{11.7} & \textbf{59.0} & \textbf{12.0} & \textbf{60.0} & \textbf{11.6} & \textbf{76.0} & \textbf{11.3} & \textbf{70.0} & \textbf{11.6} & \textbf{64.0} & \hphantom{0}\textbf{6.7} & \textbf{64.1} & \textbf{10.8} \\\midrule
ToT-DFS+Prompt & 10.0 & 48.2 & 10.5 & 50.8 & 14.5 & 45.9 & \hphantom{0}6.0 & 49.0 & 10.0 & 51.2 & \hphantom{0}8.0 & 53.3 & \hphantom{0}9.8 & 49.7 \\
\ +BTP (\emph{Ours}) & \textbf{54.5} & \textbf{11.9} & \textbf{59.5} & \textbf{12.1} & \textbf{56.5} & \textbf{11.5} & \textbf{74.5} & \textbf{11.8} & \textbf{64.0} & \textbf{11.7} & \textbf{64.0} & \hphantom{0}\textbf{6.7} & \textbf{62.2} & \textbf{11.0} \\
\bottomrule
\end{tabular}
\caption{Comparison of \textbf{Pass rate under Budget Constraint (PBC)} and \textbf{Average Cost (AC)} between the baseline methods and our proposed BTP on the ToolBench~\citep{qin2023toolllm} dataset. The experiments are conducted \textbf{with the budget constraint} $\bm{R=20}$.}
\label{tab:main_result_1}
\end{table*}

\section{Experiments}

\subsection{Setup}

\paragraph{Data Preparation.} Our experiments are mainly conducted on the ToolBench~\citep{qin2023toolllm} dataset, which is a tool learning dataset in English. We follow~\citet{qin2023toolllm} and use 6 different subsets (I1-Inst, I1-Tool, I1-Cat, I2-Inst, I2-Cat and I3-Inst) in ToolBench as the test datasets. The subset I3-Inst includes 100 user queries, while each of the other subsets contains 200 user queries.

For the past experience used for planning the tool usage, we construct a dataset using the training dataset of ToolBench. First, we use ChatGPT~\citep{openaichatgptblog} to judge whether the returned result is helpful for approximately 30k instances of tool usages in the training dataset. Then, we use these instances to train the  LongFormer-based classifier proposed in Section~\ref{sec:tool_plan} and use the classifier to classify the remaining instances in the training dataset. The total number of these two parts of instances is about 2.5M. During our experiments, all the past experience is obtained from this dataset.

\paragraph{Baselines.} We combine our proposed method with the following baseline methods and compare the combined methods with the baseline methods without such combination:
\begin{enumerate}
    \item {\bf ReAct}~\citep{yao2023react}: Tool learning is performed using a chain-of-thought process~\citep{wei2022chain}.
    \item {\bf DFSDT}~\citep{qin2023toolllm}: Tool learning is performed using a tree-based depth-first searching algorithm. This allows the LLM to traverse back when encountering a failure.
    \item {\bf ToT-DFS}~\citep{yao2023tree}: Tool learning is also performed using a depth-first searching algorithm. Different from DFSDT, ToT-DFS makes a vote across the candidate states and expands the best state at each step.
    \item {\bf ReAct+Prompt}, {\bf DFSDT+Prompt} and {\bf ToT-DFS+Prompt}: The cost of the candidate tools and the remaining budget are added to the system prompt of the corresponding method.
\end{enumerate}

\paragraph{Evaluation.} We mainly use the following evaluation metrics to evaluate the effect of budget constraint on different methods:
\begin{enumerate}
    \item {\bf Pass rate under Budget Constraint (PBC)}: The percentage of user queries which can be resolved by the LLM under the given budget constraint. When the solution given by the LLM exceeds the budget constraint, we count it as a \textbf{failure}.
    \item {\bf Average Cost (AC)}: The average cost used for all user queries in the test dataset.
\end{enumerate}

To better compare the quality of the generated solutions of our method with the baselines, we also evaluate the {\bf Pass Rate (PR)}~\cite{qin2023toolllm}, which measures the percentage of successfully resolved user queries, ignoring the budget constraint. Moreover, we also report the {\bf Rate of Failure due to Budget Constraint (RFBC)} to measure the percentage of user queries which the LLM fails to resolve because of the budget constraint.

\paragraph{Implementation Details.} We use ChatGPT~\citep{openaichatgptblog} as the backbone model for tool learning. For each user query, we retrieve 5 candidate tools using the semantic retrieval model provided by ~\citet{qin2023toolllm}. Since the ToolBench dataset does not explicitly provide the cost of the tools, we randomly set an integer in $[1,10]$ as the cost $c_i$ for each tool in the tool library. For the budget constraint, we set the value of $R$ to 20 unless explicitly specified. We set the threshold $\tau=0.15$ to filter out the useless tools. For further implementation details, please refer to Appendix~\ref{app:details}.

One might suggest that for a more accurate evaluation, a real system capable of calculating costs on the fly should be implemented. However, creating such a system involves navigating numerous real-world challenges, including the unpredictability associated with LLM behavior and network dynamics. Given that this work represents an initial exploration into budget-constrained tool learning, postponing the development of this system to future research seems reasonable. Additionally, it is crucial to clarify that the cost values $c_i$ used in the test set above are considered to be aggregate values encompassing all relevant costs, such as monetary expenses, time, and LLM usage, etc. The rationale behind this assumption and its validity is thoroughly discussed at the end of Section~\ref{sec:budget_constrained_tool_learning}.

\subsection{Main Results}

The experimental results are shown in Table~\ref{tab:main_result_1} and~\ref{tab:main_result_2}. When combining our proposed BTP with the baseline methods, the LLM can better resolve the user query within the budget constraint, regardless of which baseline method we use. Specifically, when combining BTP with the baseline methods, the average value of Pass rate under Budget Constraint (PBC) over six subsets of the ToolBench dataset increases by 9.3-53.9 points, and the Average Cost (AC) also decreases by 8.1-69.1 points. This is mainly attributed to two major aspects: First, combining BTP with the baseline methods does not significantly harm the Pass Rate (PR) even if we ignore the budget constraint. Second, the tool learning process cannot be terminated within the budget constraint for a substantial part of the user queries if we do not combine BTP with the baseline methods, while utilizing BTP can guarantee that the tool learning process can be terminated within the budget constraint.

Moreover, adding extra information to the system prompt to remind the LLM about the budget constraint is not effective for budget-constrained tool learning. For example, DFSDT+Prompt only outperforms DFSDT by 0.2 points on PBC. The average cost of DFSDT+Prompt is 73.3, which is much higher than the budget constraint $R=20$.

\subsection{Planned v.s. Actual Tool Usage}

In this section, we compare the plan given by the dynamic programming algorithm and the actual tool usage performed by the LLM. As shown in Table~\ref{tab:plan}, the average cost of the plan is 13.9 and we expect that the LLM uses the tools for 2.9 times on average on the I1-Inst subset. When BTP is combined with the baseline methods, both the actual average cost and the average number of times the LLM uses the tools are less than those given in the plan. The gap between the planned and the actual tool usage is within our expectation, since we do not actually use the tools when making the plan. Moreover, the size of the gap is reasonable, especially when we use a searching based algorithm (DFSDT or ToT-DFS) for tool learning.

\subsection{Effect of Budget Constraint $R$}

\begin{figure}[!t]
    \centering
    \includegraphics[scale=0.85]{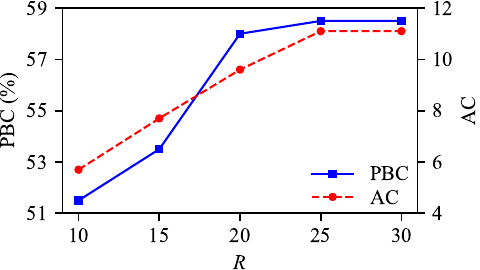}
    \caption{Effect of budget constraint $R$ on Pass rate under Budget Constraint (PBC) and Average Cost (AC). The results are evaluated on the I1-Inst subset.}
    \label{fig:fig_cost_new}
\end{figure}

To investigate how the budget constraint $R$ affects the performance of budget-constrained tool learning, we also conduct experiments on the I1-Inst subset under different budget constraints. The results are shown in Figure~\ref{fig:fig_cost_new}. On the one hand, the value of PBC increases as the budget constraint does, since a higher budget constraint gives more chances to the LLM for using the tools to resolve the user query. On the other hand, the average cost will reduce if we set a lower budget constraint, which indicates that we can adjust the cost of tool learning by modifying the budget constraint.

\begin{table}
\centering
\small
\begin{tabular}{lcc}
\toprule
\textbf{Method} & \textbf{PR}$\uparrow$ & \textbf{RFBC}$\downarrow$ \\\midrule
ReAct & \textbf{44.0} & 26.4 \\
\ +BTP (\emph{Ours}) & 43.7 & \hphantom{0}\textbf{0.0} \\\midrule
ReAct+Prompt & 44.5 & 23.9 \\
\ +BTP (\emph{Ours}) & \textbf{45.8} & \hphantom{0}\textbf{0.0} \\\midrule
DFSDT & 63.8 & 60.3 \\
\ +BTP (\emph{Ours}) & \textbf{64.5} & \hphantom{0}\textbf{0.0} \\\midrule
DFSDT+Prompt & \textbf{63.8} & 59.2 \\
\ +BTP (\emph{Ours}) & \textbf{63.8} & \hphantom{0}\textbf{0.0} \\\midrule
ToT-DFS & 61.6 & 86.3 \\
\ +BTP (\emph{Ours}) & \textbf{64.1} & \hphantom{0}\textbf{0.0} \\\midrule
ToT-DFS+Prompt & \textbf{63.1} & 86.7 \\
\ +BTP (\emph{Ours}) & 62.2 & \hphantom{0}\textbf{0.0} \\
\bottomrule
\end{tabular}
\caption{Comparison of \textbf{Pass Rate (PR)} and \textbf{Rate of Failure due to Budget Constraint (RFBC)} between the baseline methods and our proposed BTP. The experiments are conducted \textbf{with the budget constraint} $\bm{R=20}$. The detailed results are shown in Appendix~\ref{app:additional_results}.}
\label{tab:main_result_2}
\end{table}

\begin{table}
\centering
\small
\begin{tabular}{lcc}
\toprule
\textbf{Method} & \textbf{AC} & \textbf{\#Use} \\\midrule
Plan & 13.9 & 2.9 \\\midrule
ReAct+BTP & \hphantom{0}7.6 & 1.5 \\
ReAct+Prompt+BTP & \hphantom{0}8.7 & 1.7 \\
DFSDT+BTP & \hphantom{0}9.6 & 1.8 \\
DFSDT+Prompt+BTP & 10.0 & 1.9 \\
ToT-DFS+BTP & 11.7 & 2.3 \\
ToT-DFS+Prompt+BTP & 11.9 & 2.4 \\
\bottomrule
\end{tabular}
\caption{Comparison of the planned and the actual tool usage on the I1-Inst subset. ``\#Use'' represents the average number of times the LLM uses the tools for each user query.}
\label{tab:plan}
\end{table}

\subsection{Effect of Threshold $\tau$}
\begin{figure}[!t]
    \centering
    \includegraphics[scale=0.85]{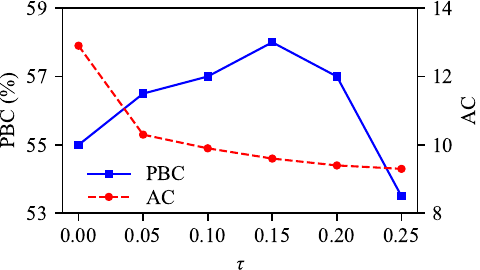}
    \caption{Effect of threshold $\tau$ on Pass rate under Budget Constraint (PBC) and Average Cost (AC). The results are evaluated on the I1-Inst subset.}
    \label{fig:fig_th_new}
\end{figure}

We introduce a threshold $\tau$ in Section~\ref{sec:tool_plan} to filter out useless tools. The effect of $\tau$ on the I1-Inst subset is shown in Figure~\ref{fig:fig_th_new}. First, if we set a higher threshold, the average cost decreases since fewer tools are available for the LLM to use. Second, the value of PBC increases when $\tau<0.15$, which indicates that setting a proper threshold can filter out the useless tools and improve the quality of the generated solutions. However, the value of PBC decreases when $\tau>0.15$, which indicates that a high threshold may incorrectly filter out some useful tools and worsen the quality of the solutions. Therefore, we set $\tau=0.15$ in our experiments.

\subsection{Effect of Blacklist Mechanism}

To validate the effectivenss of the blacklist mechanism, we conduct an extra experiment on the I1-Inst subset where the blacklist mechanism is removed. As shown in Table~\ref{tab:classifier}, removing the blacklist does not significantly affect the value of PBC. However, the average cost increases, which indicates that the blacklist mechanism can reduce the cost of repeatedly using unhelpful tools.

Moreover, to investigate how the LongFormer-based classifier affects the performance of budget-constrained tool learning, we conduct an extra experiment on the I1-Inst subset where we directly use the LLM to judge whether the returned result is helpful in the blacklist mechanism. Experimental results show that the value of PBC and the average cost do not significantly change if we use the LLM to judge the helpfulness of the returned results, even if the LLM can give the judgments more accurately than the classifier.\footnote{We randomly sample 100 returned results and use the classifier and the LLM to give the judgments, and we find the accuracy of the classifier and the LLM is 91\% and 94\%, respectively.} However, using the LongFormer-based classifier can reduce the number of times of using the LLM in practice, and thus we use the LongFormer-based classifier for our proposed BTP.

\section{Related Work}

\begin{table}
\centering
\small
\setlength{\tabcolsep}{4pt}
\begin{tabular}{lcc}
\toprule
\textbf{Blacklist} & \textbf{PBC}$\uparrow$ & \textbf{AC}$\downarrow$ \\\midrule
None & 57.5 & 10.2 \\
ChatGPT & \textbf{58.0} & \hphantom{0}9.7 \\\midrule
LongFormer & \textbf{58.0} & \textbf{\hphantom{0}9.6} \\
\bottomrule
\end{tabular}
\caption{Effect of blacklist mechanism on I1-Inst.}
\label{tab:classifier}
\end{table}

Tool learning aims to extend the capability of LLMs using external tools~\citep{schick2023toolformer,yao2023react,qin2023tool,qin2023toolllm}. With the aid of external tools, LLMs may be capable of obtaining recent information~\citep{liu2023webglm}, solving difficult math problems~\citep{cobbe2021training,gao2023pal}, handling multimodal information~\citep{huang2023audiogpt,lu2023chameleon} or solving domain-specific problems~\citep{jin2023genegpt}.

The basic approach of tool learning is to add the documentation of the tool to the input context of the LLM and then let the LLM use the tool to resolve the user query~\citep{shen2023hugginggpt,hsieh2023tool,qin2023toolllm}. To handle complex user queries which require multiple steps to resolve, \citet{yao2023react} propose ReAct, which performs tool learning in a chain-of-thought~\citep{wei2022chain} process. To increase the probability of successfully resolving the user query, \citet{qin2023toolllm} propose DFSDT, a tool learning algorithm which is based on a tree-based depth-first searching algorithm and allows the LLM to traverse back after a failure. Similarly, \citet{zhuang2023toolchain} also propose a tree-based algorithm for tool learning, which is based on the A* searching algorithm. Moreover, \citet{gao2023assistgpt} and~\citet{gao2023clova} add a reflection mechanism to tool learning, which allows the LLM to retry after a failure.

However, the aforementioned studies focus on how to better resolve complex user queries, but do not take the budget constraint into consideration. In contrast, our proposed method aims to improve the performance of tool learning within the budget constraint of tool usage.

This work is also highly related to~\citet{kim2023llm}, who use different tools in parallel to improve the efficiency of tool learning. However, their work mainly focuses on time efficiency rather than budget constraint.

\section{Conclusion}

In this work, we propose BTP, a novel method for budget-constrained tool learning. The core idea of BTP is to find a preferable plan under the budget constraint before using the tools. We also introduce a blacklist mechanism to further reduce the cost of tool learning. Experimental results show that BTP can be combined with different tool learning methods and improve the performance of budget-constrained tool learning.

\section*{Limitations}

In this work, we propose a simplified yet effective approximation to model budget-constrained tool learning. As discussed in Section~\ref{sec:budget_constrained_tool_learning}, our approach does not account for costs incurred by external factors, which are challenging to estimate based on previous experience of tool learning. Exploring a more comprehensive mathematical model that includes these costs is an avenue for future research. Furthermore, our experiments are carried out in a simulated setting, where the costs associated with the tools are not based on real-world data but are simulated instead. Creating similar conditions in a real-world context to accurately estimate tool costs poses numerous challenges. Therefore, developing techniques to estimate real-world tool costs is an objective for future research.

\section*{Acknowledgments}

This work is supported by the National Natural Science Foundation of China (No. 61925601, 62276152). We thank Zonghan Yang for providing helpful suggestions about the idea of this work. We also appreciate the anonymous reviewers for their valuable comments.

\bibliography{acl_latex}

\clearpage

\appendix

\section{Appendix}

\subsection{Proof of Approximation of $c_i$ and $R$}
\label{app:proof_approximation}

In this proof, we suppose that the tool learning process consists of a reasonable number of steps, and we use $N$ to represent the number of steps. For any $\varepsilon>0$, we set $\lambda$ as:
\begin{equation}
    \lambda=\frac{N+1}{\varepsilon}
\end{equation}

Then, we use $\tilde{c}_i$ and $\tilde{R}$ to denote the approximate value of $c_i$ and $R$. $\tilde{c}_i$ and $\tilde{R}$ are defined as below:
\begin{align}
\tilde{c}_i &= \frac{\lceil\lambda c_i\rceil}{\lambda}, \\
\tilde{R} &= \frac{\lfloor\lambda R\rfloor}{\lambda}.
\end{align}

Based on the equations above, $\tilde{c}_i$ and $\tilde{R}$ satisfy $\lambda\tilde{c}_i \in \mathbb{Z}$ and $\lambda\tilde{R} \in \mathbb{Z}$.

Besides, $\tilde{c}_i$ and $\tilde{R}$ satisfy
\begin{equation}
\begin{aligned}
R-\tilde{R}&=\frac{\lambda R-\lfloor\lambda R\rfloor}{\lambda} \\
&< \frac{1}{\lambda} \\
&= \frac{\varepsilon}{N+1},
\end{aligned}
\end{equation}
\begin{equation}
\begin{aligned}
\tilde{c}_i-c_i&=\frac{\lceil\lambda c_i\rceil-\lambda c_i}{\lambda} \\
&< \frac{1}{\lambda} \\
&= \frac{\varepsilon}{N+1}.
\end{aligned}
\end{equation}

Thus, the approximate remaining budget $\tilde{R}-\sum\limits_{j=1}^{N} \tilde{c}_{a_j}$ satisfies
\begin{equation}
\begin{aligned}
&\quad\ \tilde{R}-\sum\limits_{j=1}^{N} \tilde{c}_{a_j} \\
&> R-\frac{\varepsilon}{N+1}-\sum\limits_{j=1}^{N}\left(c_{a_j}+\frac{\varepsilon}{N+1}\right) \\
&=R-\sum\limits_{j=1}^{N}c_{a_j}-\varepsilon.
\end{aligned}
\end{equation}

Moreover, since $\tilde{R} = \frac{\lfloor\lambda R\rfloor}{\lambda} \leq R$ and $\tilde{c}_i = \frac{\lceil\lambda c_i\rceil}{\lambda} \geq c_i$, $\tilde{R}-\sum\limits_{j=1}^{N} \tilde{c}_{a_j}$ also satisfies
\begin{equation}
\tilde{R}-\sum\limits_{j=1}^{N}\tilde{c}_{a_j} \leq R-\sum\limits_{j=1}^{N}c_{a_j}.
\end{equation}

Thus, we conclude that
\begin{equation}
\left|\left(\tilde{R}-\sum\limits_{j=1}^{N}\tilde{c}_{a_j}\right)-\left(R-\sum\limits_{j=1}^{N}c_{a_j}\right)\right| < \varepsilon.
\end{equation}

This indicates that if $c_i$ and $R$ are real numbers, we can use $\lambda\tilde{c}_i$ and  $\lambda\tilde{R}$ as their integral approximation. If the number of the total steps taken for tool learning $N$ is reasonable, such approximation can achieve sufficiently high accuracy if $\varepsilon$ is small enough.

\subsection{Further Implementation Details}
\label{app:details}

The training process of the LongFormer-based classification model is implemented on top of the Transformers~\citep{wolf2020transformers} library. The model is initialized with \texttt{longformer-base-4096}\footnote{https://huggingface.co/allenai/longformer-base-4096}, which has approximately 149M parameters. The model is trained for 10 epochs with the learning rate of $10^{-5}$. The batch size is set to 8. The model is trained on a single NVIDIA GeForce RTX 3090 GPU for 5.5 hours.

\begin{figure*}
\begin{center}
\includegraphics[scale=0.75]{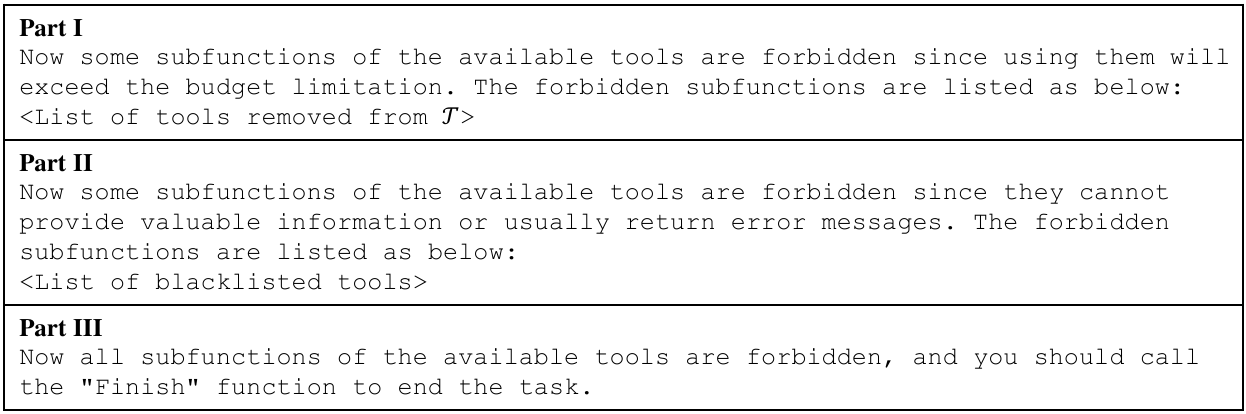} 
\caption{The extra content which may be appended to the system prompt during the tool learning process.}
\label{fig:extra_prompt}
\end{center}
\end{figure*}

Moreover, the system prompt used during in the tool learning process is constructed based on the prompt for solution path annotation introduced in~\citet{qin2023toolllm}. We add extra content presented in Figure~\ref{fig:extra_prompt} in the system prompt if some of the tools are forbidden to use. Specifically, we append the prompt {\bf Part I} to the system prompt if at least one candidate tool is removed from $\mathcal{T}$. We append the prompt {\bf Part II} to the system prompt if at least one candidate tool is blacklisted. We append the prompt {\bf Part III} to the system prompt if all the candidate tools are forbidden to use (i.e. removed from $\mathcal{T}$ or blacklisted).

\subsection{Detailed Experimental Results of Table~\ref{tab:main_result_2}}
\label{app:additional_results}

In this section, we report the detailed experimental results of Table~\ref{tab:main_result_2} in Table~\ref{tab:main_result_pass_rate} and~\ref{tab:main_result_rfbc}.

\subsection{Case Study}

\begin{figure*}[t]
\begin{center}
\includegraphics[scale=0.74]{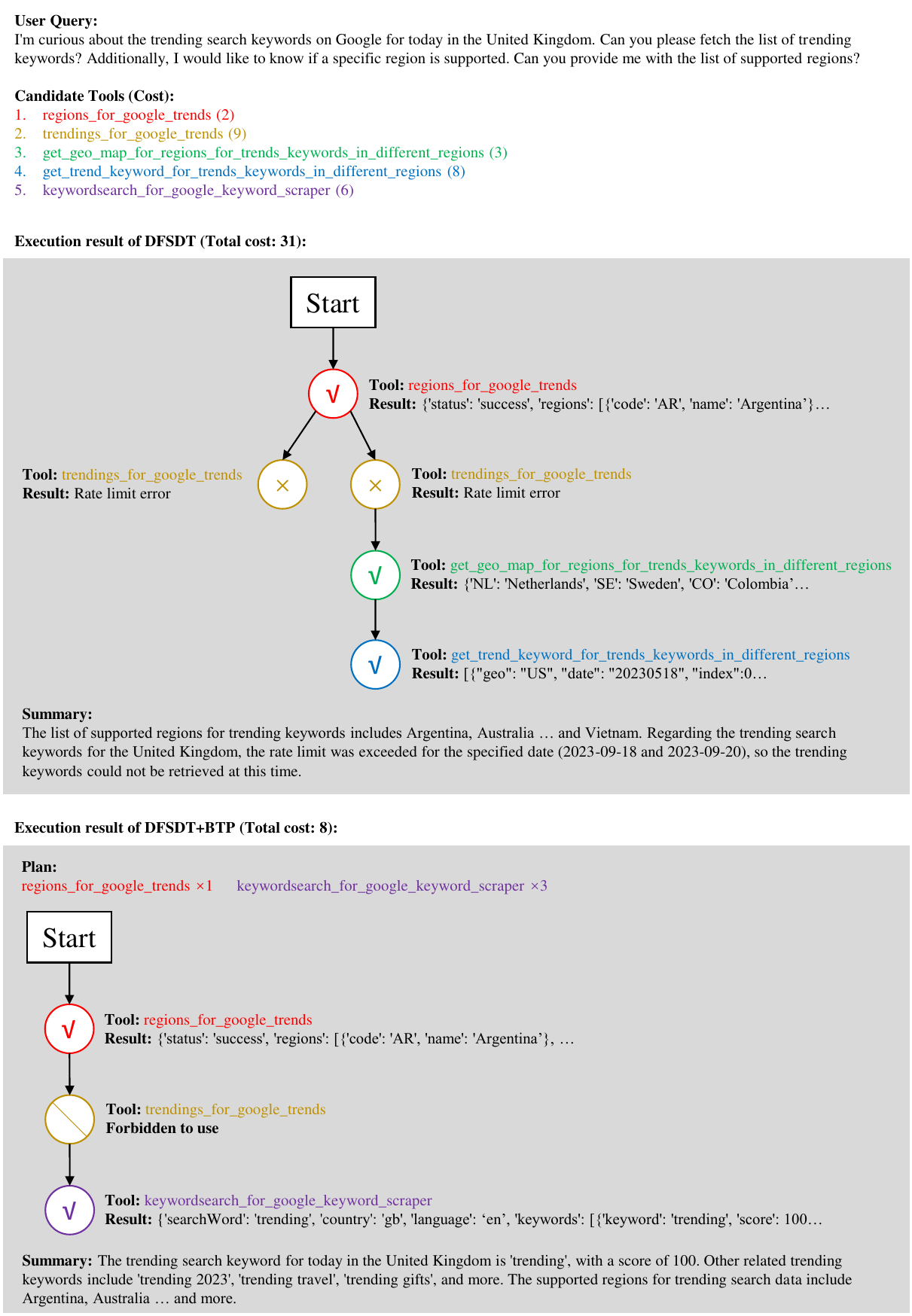} 
\caption{Case study on the I1-Inst subset.}
\label{fig:casestudy}
\end{center}
\end{figure*}

In this section, we conduct a case study on the I1-Inst subset. As shown in Figure~\ref{fig:casestudy}, the baseline method DFSDT repeatedly uses the tool 2 which always returns error messages. DFSDT also uses the tools 3 and 4 which return repetitive or irrelevant results. As the result, DFSDT finishes the tool learning process with the total cost of 31. It only finds the list of supported regions but fails to find the trending keywords in the United Kingdom.

When combining our proposed BTP with DFSDT, the LLM uses the tool 5 within the plan and successfully finds the trending keywords in the United Kingdom. Thus, DFSDT+BTP successfully resolves the user query with the total cost of 8. This indicates that our proposed BTP can help the LLM resolve the user query within the budget constraint.

\subsection{Licenses of Tools, Models and Datasets}

The licenses of the Transformers~\citep{wolf2020transformers} library, the LongFormer~\citep{beltagy2020longformer} model and the ToolBench~\citep{qin2023toolllm} dataset are Apache-2.0.

\begin{table*}
\centering
\small
\begin{tabular}{lccccccc}
\toprule
\textbf{Method} & \textbf{I1-Inst} & \textbf{I1-Tool} & \textbf{I1-Cat} & \textbf{I2-Inst} & \textbf{I2-Cat} & \textbf{I3-Inst} & \textbf{Average} \\\midrule
ReAct & 47.0 & 51.5 & 45.0 & 45.0 & 47.5 & 28.0 & 44.0 \\
\ +BTP (\emph{Ours}) & 43.5 & 48.0 & 49.5 & 41.5 & 49.5 & 30.0 & 43.7 \\\midrule
ReAct+Prompt & 43.5 & 52.5 & 48.0 & 44.5 & 49.5 & 29.0 & 44.5 \\
\ +BTP (\emph{Ours}) & 47.0 & 44.0 & 50.5 & 47.0 & 54.0 & 32.0 & 45.8 \\\midrule
DFSDT & 56.0 & 62.0 & 56.5 & 76.0 & 68.5 & 64.0 & 63.8 \\
\ +BTP (\emph{Ours}) & 58.0 & 58.0 & 63.5 & 77.0 & 66.5 & 64.0 & 64.5 \\\midrule
DFSDT+Prompt & 54.0 & 63.5 & 60.0 & 76.0 & 65.5 & 64.0 & 63.8 \\
\ +BTP (\emph{Ours}) & 55.0 & 58.0 & 63.0 & 77.5 & 65.5 & 64.0 & 63.8 \\\midrule
ToT-DFS & 53.5 & 60.0 & 54.5 & 73.5 & 65.0 & 63.0 & 61.6 \\
\ +BTP (\emph{Ours}) & 55.5 & 59.0 & 60.0 & 76.0 & 70.0 & 64.0 & 64.1 \\\midrule
ToT-DFS+Prompt & 54.5 & 61.5 & 58.5 & 74.5 & 66.5 & 63.0 & 63.1 \\
\ +BTP (\emph{Ours}) & 54.5 & 59.5 & 56.5 & 74.5 & 64.0 & 64.0 & 62.2 \\
\bottomrule
\end{tabular}
\caption{Comparison of \textbf{Pass Rate (PR)} between the baseline methods and our proposed BTP on the ToolBench~\citep{qin2023toolllm} dataset. The experiments are conducted \textbf{with the budget constraint} $\bm{R=20}$.}
\label{tab:main_result_pass_rate}
\end{table*}

\begin{table*}
\centering
\small
\begin{tabular}{lccccccc}
\toprule
\textbf{Method} & \textbf{I1-Inst} & \textbf{I1-Tool} & \textbf{I1-Cat} & \textbf{I2-Inst} & \textbf{I2-Cat} & \textbf{I3-Inst} & \textbf{Average} \\\midrule
ReAct & 27.0 & 30.0 & 22.5 & 28.5 & 31.5 & 19.0 & 26.4 \\
ReAct+Prompt & 27.0 & 30.0 & 17.5 & 20.5 & 24.5 & 24.0 & 23.9 \\
DFSDT & 56.0 & 61.0 & 53.5 & 62.0 & 63.5 & 66.0 & 60.3 \\
DFSDT+Prompt & 54.5 & 62.5 & 53.0 & 57.0 & 57.0 & 71.0 & 59.2 \\
ToT-DFS & 85.5 & 84.5 & 87.0 & 92.5 & 87.0 & 81.0 & 86.3 \\
ToT-DFS+Prompt & 85.0 & 86.5 & 81.0 & 91.0 & 86.5 & 90.0 & 86.7 \\
\bottomrule
\end{tabular}
\caption{\textbf{Rate of Failure due to Budget Constraint (RFBC)} of the baseline methods on the ToolBench~\citep{qin2023toolllm} dataset. The experiments are conducted \textbf{with the budget constraint} $\bm{R=20}$. For our proposed BTP, the value of RFBC is consistently equal to 0 since the plan made in Algorithm~\ref{alg:dp} guarantees that the tool learning process can be finished within the budget constraint.}
\label{tab:main_result_rfbc}
\end{table*}

\end{CJK}
\end{document}